\documentclass[runningheads]{llncs}
\usepackage[T1]{fontenc}
\usepackage{array}
\usepackage{booktabs}
\usepackage{multirow}
\usepackage[table,xcdraw]{xcolor}
\usepackage{marvosym}
\usepackage{graphicx}
\usepackage{amsmath}
\usepackage[ruled]{algorithm2e}
\makeatletter
\def\thanks{\protected@xdef\@thanks{\@thanks\protect\footnotetext{\dag\ Equal contributions}}}
\makeatother

\begin{document}
\title{BoxPolyp: Boost Generalized Polyp Segmentation using Extra Coarse Bounding Box Annotations}


\author{
    Jun Wei\inst{1,2,3,\dag},                  
    Yiwen Hu\inst{1,2,3,6,\thanks{\dag}},
    Guanbin Li\inst{7},\\                
    Shuguang Cui\inst{1,2,3},            
    S.Kevin Zhou\inst{1,4,5}             
    Zhen Li\inst{1,2,3,\textrm{\Letter}} 
} 

\institute{
    $^1$ School of Science and Engineering, The Chinese University of Hong Kong(Shenzhen) \\
    $^2$ Shenzhen Research Institute of Big Data \\
    $^3$ The Future Network of Intelligence Institute \\
    $^4$ School of Biomedical Engineering \& Suzhou Institute for Advanced Research, University of Science and Technology of China, Suzhou, China\\
    $^5$ Institute of Computing Technology, Chinese Academy of Sciences, Beijing, China\\
    $^6$ Institute of Urology, The Third Affiliated Hospital of Shenzhen University(Luohu Hospital Group), Shenzhen, China.\\
    $^7$ School of Computer Science and Engineering, Sun Yat-Sen University, China
    \email{lizhen@cuhk.edu.cn}
}

\maketitle


\begin{abstract}
Accurate polyp segmentation is of great importance for colorectal cancer diagnosis and treatment. 
However, due to the high cost of producing accurate mask annotations, existing polyp segmentation methods suffer from severe data shortage and impaired model generalization. Reversely, coarse polyp bounding box annotations are more accessible.
Thus, in this paper, we propose a boosted {\textbf{BoxPolyp}} model to make full use of both accurate mask and extra coarse box annotations.
In practice, box annotations are applied to alleviate the over-fitting issue of previous polyp segmentation models, which generate fine-grained polyp area through the iterative boosted segmentation model. 
To achieve this goal, a fusion filter sampling (FFS) module is firstly proposed to generate pixel-wise pseudo labels from box annotations with less noise, leading to significant performance improvements.
Besides, considering the appearance consistency of the same polyp, an image consistency (IC) loss is designed. 
Such IC loss explicitly narrows the distance between features extracted by two different networks, which improves the robustness of the model.
Note that our BoxPolyp is a plug-and-play model, which can be merged into any appealing backbone.
Quantitative and qualitative experimental results on five challenging benchmarks confirm that our proposed model outperforms previous state-of-the-art methods by a large margin.
\end{abstract}


\section{Introduction}
\label{introduction}
Colorectal Cancer (CRC) is one of the leading causes of malignant tumors death worldwide. As the precursor of CRC, colorectal polyps are the driver of CRC morbidity and mortality. Therefore, accurate polyp segmentation and diagnosis are of great significance to the survival of patients. Thanks to the evolution of computer technology, massive polyp segmentation models~\cite{akbari2018polyp,brandao2017fully,dong2021polyp,fan2020pranet,fang2019selective,ronneberger2015u,tajbakhsh2015automated,wei2021shallow,zhang2020adaptive,zhang2018road,zhou2018unet++} have been proposed and achieved remarkable performance.

However, these models are always plagued by data shortages and suffer from severe over-fitting issue. Since the popularity of U-Net~\cite{ronneberger2015u} and FCN~\cite{long2015fully}, most of polyp segmentation models~\cite{fan2020pranet,wei2021shallow} are based on convolutional neural networks (CNNs), which outperform traditional handcrafted ones but are data hungry. 
Unfortunately, accurate labeling of polyp masks is time-consuming and laborious, requiring pixel-by-pixel operation. 
Therefore, existing polyp segmentation datasets are relatively small. In particular, the widely adopted polyp training set~\cite{fan2020pranet,wei2021shallow} contains only 1,451 images, far from enough to feed a large capacity CNN model.
Thus, models trained on this dataset exhibit the unstable performance and are sensitive to noise, which hampers the practical clinical usage.

\begin{figure}[t]
  \centering
  \includegraphics[scale=0.3]{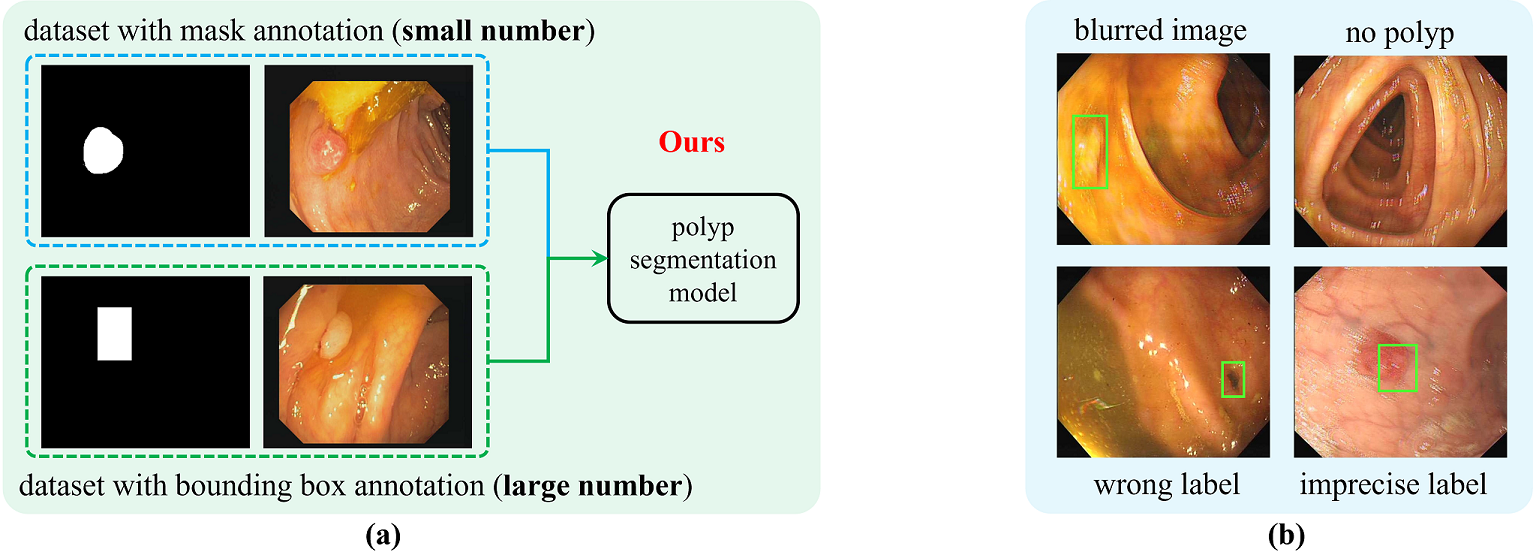}
  \caption{(a) Our proposed polyp segmentation model using both accurate mask annotations and coarse bounding box ones. (b) Common annotation issues of the polyp detection dataset LDPolypVideo~\cite{ma2021ldpolypvideo}. The second row shows the noisy annotations.}
  \label{fig:teaser}
\end{figure}

To provide effective clinical assistance, a generalized polyp segmentation model is urgently needed. 
In this paper, we struggle to achieve this goal using extra coarse bounding box annotations to expand the small segmentation dataset.
Specifically, a large open-released polyp detection dataset LDPolypVideo~\cite{ma2021ldpolypvideo} is adopted, which consists of 160 polyp video clips with 40,266 frames. Though all these images are labeled with only coarse bounding box annotations, they provide sufficient polyp appearance information and are much cheaper.
%
%
Fig.~\ref{fig:teaser}(a) shows the core idea of our method. Tranined with a few images with mask annotations and a lot of images with bounding box annotations, a more generalized polyp segmentation model is achieved.
However, directly applying the bounding box annotations of LDPolypVideo is suboptimal. Because the bounding box area contains many background pixels. Taking the bounding box area as polyp mask will bring a lot of noise. Besides, as shown in Fig.~\ref{fig:teaser}(b), LDPolypVideo contains many blurred images, images with no polyps, wrong labels and imprecise labels, which also will mislead the model training.

To make the most of the good parts of LDPolypVideo annotations and reduce the bad parts, we propose the novel BoxPolyp model which mainly consists of two modules: fusion filter sampling and image consistency loss. 
In practice, fusion filter sampling (FFS) aims to generate pseudo labels for high-confidence regions of each image in LDPolypVideo. By combining the raw bounding box annotations and the predicted masks (derived from the model trained on a small polyp segmentation dataset), FFS efficiently produces pixel-wise pseudo masks for deterministic regions. For uncertain regions, pseudo masks are inaccurate and therefore discarded.
%
However, these discarded regions also contain valuable information. To fully explore these regions, we propose the image consistency (IC) loss instead of generating pseudo masks. IC loss applies two different networks to extract features from the same image and explicitly reduces the distance between features of the uncertain regions. By forcing feature alignment, our model could learn robust polyp feature representations, requiring no mask annotations.

In summary, our contributions are three-folds: (1) We are the first to boost a generalized polyp segmentation model through extra bounding box annotations. (2) We propose the fusion filter sampling to generate pseudo masks with less noise and design the image consistency loss to enhance the feature robustness of uncertain regions. (3) Our proposed BoxPolyp is a plug-and-play model, which can largely enhance polyp segmentation performance using different backbones.


\section{Related Work}
Traditional polyp segmentation models~\cite{tajbakhsh2015automated,zhou2020review} are mostly based on low-level features ({\it i.e.}, color, texture and boundary). But limited by the poor semantics, these models fail when dealing with complex scenarios.
Recently, fully convolutional networks (FCN)~\cite{long2015fully} have been widely adopted for polyp segmentation and make great progress. 
For example, U-Net~\cite{ronneberger2015u}, U-Net++~\cite{zhou2018unet++} and ResUNet++~\cite{jha2019resunet++} use the encoder-decoder architecture to handle the segmentation tasks, which has become the standard paradigm for subsequent works. However, the polyp boundaries are not well handled by these methods. 
Afterwards, PsiNet~\cite{murugesan2019psi}, LODNet~\cite{cheng2021learnable}, PraNet~\cite{fan2020pranet}, MSNet~\cite{zhao2021automatic} and SFANet~\cite{fang2019selective} force the model to learn the feature differences, which greatly enhances the model's perception for polyp boundaries and achieve the promising results. 

Besides, ACSNet~\cite{zhang2020adaptive}, HRENet~\cite{shen2021hrenet} and CCBANet~\cite{nguyen2021ccbanet} pay more attention to context information. By adaptively aggregating multi-scale contexts, the ambiguity of local features will be reduced, thus leading to highly confident predictions.
Unlike the above methods, SANet~\cite{wei2021shallow} deals with the polyp segmentation task in terms of data distribution. By eliminating the color bias of the image, SANet achieves robust performance gains in different scenarios. 
Furthermore, with the success of transformer in image processing, researchers start working on the long-distance dependency. For instance, PNSNet~\cite{ji2021progressively} uses a self-attention block to mine the temporal and spatial relations in polyp videos. Polyp-Pvt~\cite{dong2021polyp} directly introduces a transformer encoder to replace the widely used CNN backbones. Differently, Transfuse~\cite{zhang2021transfuse} combines both CNN and transformer to extract spatial correlation and global context. All these methods have achieved remarkable performance. But limited by training set, these models suffer from the over-fitting issue. Therefore, we propose to use the cheap bounding box annotations to boost a generalized polyp segmentation model.

\section{Method}
\begin{figure}[t]
  \centering
  \includegraphics[scale=0.33]{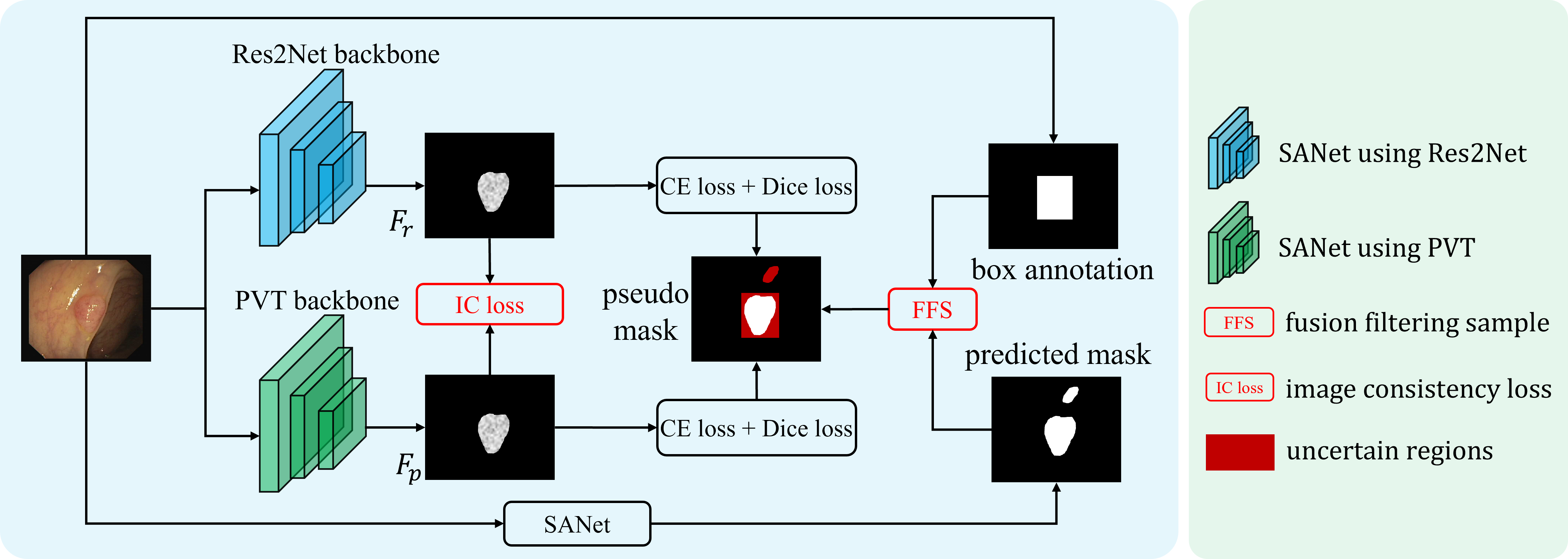}
  \caption{The pipeline for our proposed BoxPolyp model. First, a SANet~\cite{wei2021shallow} trained on the small polyp segmentation dataset is used to predict the pixel-wise mask for each box-annotated image. Then, a FFS module combines the predicted mask and the box annotation to get the deterministic regions as pseudo labels. For regions of uncertainty, we propose the IC loss to reduce the distance between features extracted from two different backbones({\it i.e.,} Res2Net~\cite{GaoCZZYT21} and PVT~\cite{wang2021pvtv2})}.
  \label{fig:framework}
\end{figure}
Fig.~\ref{fig:framework} depicts the whole framework of the proposed BoxPolyp segmentation model, consisting of two parts: fusion filter sampling and image consistency loss. Without special instructions, we use SANet~\cite{wei2021shallow} as our baseline model. 

\subsection{Fusion Filter Sampling}
We integrate polyp detection dataset ({\it i.e.,} LDPolypVideo~\cite{ma2021ldpolypvideo}) to enhance the polyp segmentation model. But LDPolypVideo is flawed in two ways. First, as shown in Fig.~\ref{fig:teaser}(b), there exists many wrongs and imprecise labels in LDPolypVideo, bringing noise for supervision. Second, bounding box annotations only provide coarse polyp contours and some background pixels are also included. Directly taking bounding box masks as pseudo labels will mislead the model. To solve the above issues, we propose fusion filter sampling (FFS) to generate pseudo masks with less noise interference. 

\begin{figure}[t]
    \centering
    \includegraphics[scale=0.205]{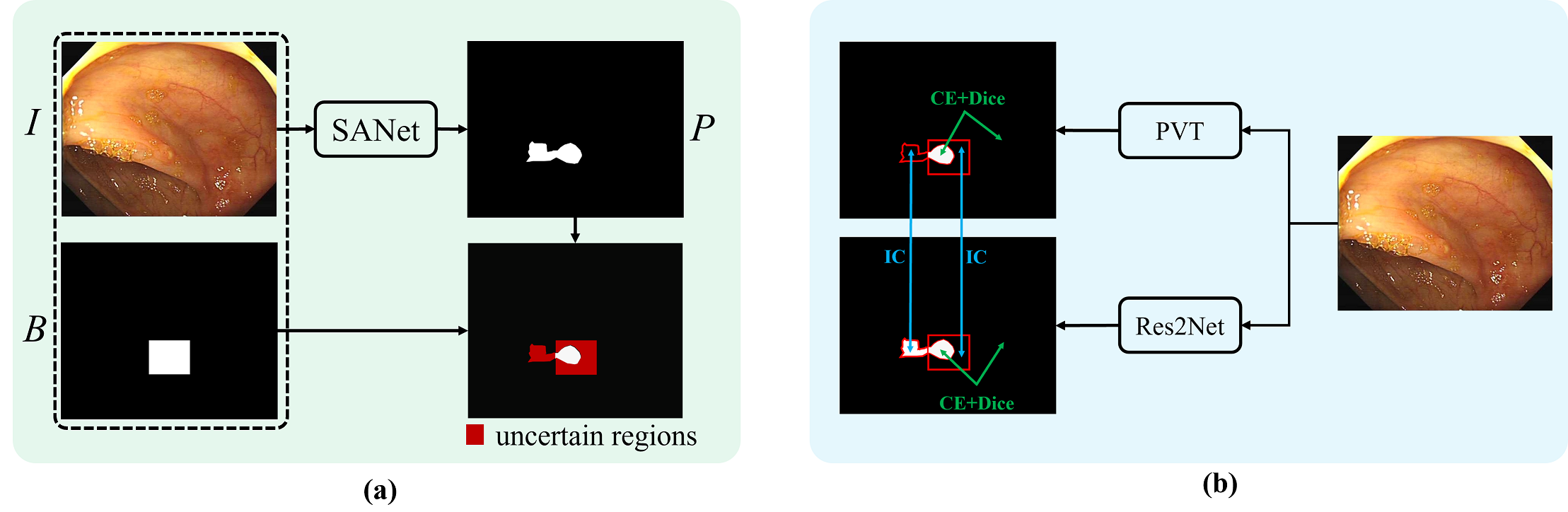}
    \caption{(a) Refined pseudo mask generation using fusion filter sampling module, which consists of foreground, background and uncertain regions. (b) Different supervision for regions of certainty and regions of uncertainty.}
    \label{fig:pseudo}
\end{figure}

Specifically, FFS filters out noise through object-level bounding box annotations and pixel-level pseudo masks, where object-level annotations weed out mislabeled or hard images and pixel-level masks filter out the background pixels in bounding box regions. 
For the object-level operation, given an image $I$, we first convert its bounding box annotations into a binary mask $B$, as shown in Fig.~\ref{fig:pseudo}(a). Meanwhile, a pre-trained SANet~\cite{wei2021shallow} model (trained on small polyp segmentation dataset) is applied to get a coarse prediction $P$ for $I$. 
Intuitively, if there is a big difference between $B$ and $P$, $I$ may be a hard sample or a mislabeled sample. In either case, $I$ will be filtered out and not involved in the model training. Thus, the issues shown in Fig.~\ref{fig:teaser}(b) will be alleviated.
In practice, we choose Dice $d=\frac{2BP}{B+P}$ to measure the difference between $B$ and $P$. Only images with $d>0.7$ will be selected out to minimize the impact of object-level wrong annotations.
For the pixel-level operation, we combine the complementarity of $B$ and $P$ to refine the pseudo masks. Specifically, we choose pixels where both $B$ and $P$ are equal to 1 as the foreground $F$. Namely, $F=B \cap P$. Similarly, only pixels where both $B$ and $P$ are equal to 0 will be regarded as the background $K$. Namely, $K=(1-B) \cap (1-P)$. Other pixels belong to the uncertain regions, as shown in Fig.~\ref{fig:pseudo}(a). During training, only $F$ and $K$ are involved in supervision, while uncertain regions are dealt with IC loss (described in Sec.~\ref{IC_loss}). Through FFS, we maximize the utilization of bounding box annotations and minimize the potential noise interference.

\subsection{Image Consistency (IC) Loss}
\label{IC_loss}
By combining bounding box annotations and predicted masks, FFS module obtains deterministic foreground and background regions for supervision, as shown in Fig.~\ref{fig:pseudo}(a). However, the regions of uncertainty are not supervised during training. Because no matter the box mask or the predicted mask is used as the pseudo label, it will bring a lot of noise which is harmful to the model generalization. In view of this, we propose the image consistency loss which mines supervisory information from the relationship between images, instead of pseudo labels.

Specifically, for each polyp image, we send it to two SANet models but with different backbone networks ({\it i.e,.} Res2Net~\cite{GaoCZZYT21} and PVT~\cite{wang2021pvtv2}), as shown in Fig.~\ref{fig:framework}. Due to the different architectures ({\it i.e,.} CNN and Transformer), the features $F_r$ and $F_p$ extracted by Res2Net and PVT present different characteristics. Meanwhile, $F_r$ and $F_p$ come from the same image. They should have similar appearance. To bring the supervision for regions of uncertainty, we propose the IC loss to explicitly reduce the distance between $F_r$ and $F_p$, as shown in Eq.~\ref{eq:ic}.
\begin{align}
\mathcal{L}_{IC} = \frac{\sum_{i,j}(F_r^{i,j}-F_p^{i,j})^2 \cdot U^{i,j}}{\sum_{i,j}U^{i,j}} \label{eq:ic}
\end{align}
where $i$ and $j$ are the pixel indexes of polyp regions, $U$ represents the mask of uncertain regions. Thus, IC loss focuses on regions without labels. Supervised by the IC loss, our model outputs more consistent predictions and greatly reduces the over-fitting risk.

\subsection{Loss Function}
Following previous methods~\cite{wei2021shallow,fan2020pranet}, binary cross entropy $\mathcal{L}_{BCE}$ and Dice loss $\mathcal{L}_{Dice}$ are adopted. Besides, the proposed $\mathcal{L}_{IC}$ is also involved in the total loss, as shown in Eq.~\ref{eq:loss}.
\begin{equation}
\label{eq:loss}
 \mathcal{L}_{total} = \mathcal{L}_{BCE} + \mathcal{L}_{Dice} + \mathcal{L}_{IC}
\end{equation}

\section{Experiments}
\subsection{Datasets and Training Settings}
Five widely used polyp segmentation datasets are adopted to evaluate the model performance, including Kvasir~\cite{jha2020kvasir}, CVC-ClinicDB~\cite{bernal2015wm}, CVC-ColonDB~\cite{bernal2012towards},  EndoScene~\cite{vazquez2017benchmark} and ETIS~\cite{silva2014toward}. For the comparability, we follow the same dataset partition as~\cite{fan2020pranet}. Besides, nine state-of-the-art methods are used for comparison, namely U-Net~\cite{ronneberger2015u}, U-Net++~\cite{zhou2018unet++}, ResUNet~\cite{zhang2018road}, ResUNet++~\cite{jha2019resunet++}, SFA~\cite{fang2019selective}, PraNet~\cite{fan2020pranet}, SANet~\cite{wei2021shallow}, MSNet~\cite{zhao2021automatic} and Polyp-Pvt~\cite{dong2021polyp}. Pytorch is used to implement our BoxPolyp model. All input images are uniformly resized to 352×352. For data augmentation, random flip, random rotation and multi-scale training are adopted. The whole network is trained in an end-to-end way with a AdamW optimizer. Initial learning rate and batch size are set to 1e-4 and 16, respectively. We train the entire model for 80 epochs.

\subsection{Quantitative Comparison}
\begin{table}[t]
\centering
\caption{Performance comparison with different polyp segmentation models. The \textcolor{red}{red} column represents the weighted average (wAVG) performance of different testing datasets. Below the dataset name is the image number of each dataset.}
\label{tab:performace}
\renewcommand\arraystretch{1.2}
\renewcommand\tabcolsep{2.5pt}
\begin{tabular}{lcccccccccccc}
    \hline
        & \multicolumn{2}{c}{ColonDB} & \multicolumn{2}{c}{Kvasir} & \multicolumn{2}{c}{ClinicDB} & \multicolumn{2}{c}{EndoScene} & \multicolumn{2}{c}{ETIS} & \multicolumn{2}{c}{\textcolor{red}{wAVG}} \\
        & \multicolumn{2}{c}{380} & \multicolumn{2}{c}{100} & \multicolumn{2}{c}{62} & \multicolumn{2}{c}{60} & \multicolumn{2}{c}{196} & \multicolumn{2}{c}{\textcolor{red}{798}} \\
    \multirow{-3}{*}{Methods}   & Dice & IoU  & Dice & IoU  & Dice & IoU  & Dice & IoU  & Dice & IoU  & \textcolor{red}{Dice} & \textcolor{red}{IoU} \\ 
    \hline
    U-Net                       & .512 & .444 & .818 & .746 & .823 & .750 & .710 & .627 & .398 & .335 & \textcolor{red}{.561} & \textcolor{red}{.493}\\
    U-Net++                     & .483 & .410 & .821 & .743 & .794 & .729 & .707 & .624 & .401 & .344 & \textcolor{red}{.546} & \textcolor{red}{.476}\\
    ResUNet                     & -    & -    & .791 & -    & .779 & -    & -    & -    & -    & -    & -    & -\\
    ResUNet++                   & -    & -    & .813 & .793 & .796 & .796 & -    & -    & -    & -    & -    & - \\
    SFA                         & .469 & .347 & .723 & .611 & .700 & .607 & .467 & .329 & .297 & .217 & \textcolor{red}{.476} & \textcolor{red}{.367}\\
    PraNet                      & .712 & .640 & .898 & .840 & .899 & .849 & .871 & .797 & .628 & .567 & \textcolor{red}{.741} & \textcolor{red}{.675}\\
    MSNet                       & .751 & .671 & .905 & .849 & .918 & .869 & .865 & .799 & .723 & .652 & \textcolor{red}{.785} & \textcolor{red}{.714}\\
    \hline
    SANet                       & .753 & .670 & .904 & .847 & .916 & .859 & .888 & .815 & .750 & .654 & \textcolor{red}{.794} & \textcolor{red}{.714}\\  
    \textbf{Ours-Res2Net}       & .820 & .741 & .910 & .857 & .904 & .849 & .903 & .835 & .829 & .742 & \textbf{\textcolor{red}{.846}} & \textbf{\textcolor{red}{.771}}\\
    \hline
    Polyp-Pvt                   & .808 & .727 & .917 & .864 & .937 & .889 & .900 & .833 & .787 & .706 & \textcolor{red}{.833} & \textcolor{red}{.760}\\  
    \textbf{Ours-Pvt}           & .819 & .739 & .918 & .868 & .918 & .868 & .906 & .840 & .842 & .755 & \textbf{\textcolor{red}{.851}} & \textbf{\textcolor{red}{.776}}\\
    \hline
\end{tabular}
\end{table}

\begin{figure}[t]
  \centering
  \includegraphics[scale=0.5]{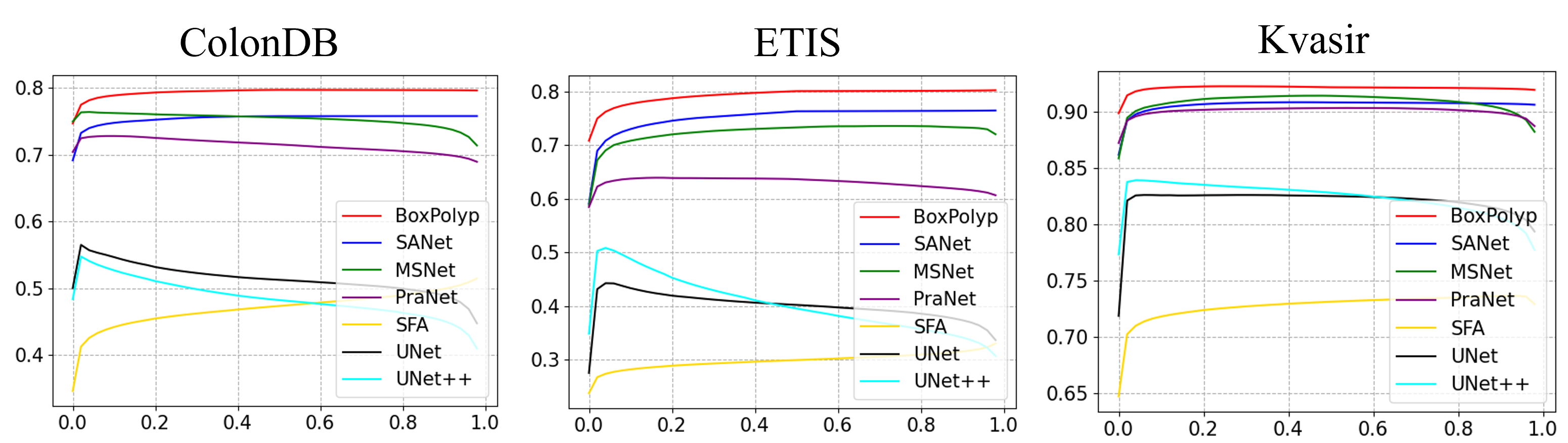}
  \caption{Dice curves under different thresholds on three polyp datasets.}
  \label{fig:dicecurve}
\end{figure}

To prove the effectiveness of the proposed BoxPolyp, nine state-of-the-art models are used for comparison, as shown in Table~\ref{tab:performace}. BoxPolyp surpasses previous methods by a large margin on the weighted average (wAVG) performace of five datasets, demonstrating the superior performance of the proposed methods. In addition, Fig.~\ref{fig:dicecurve} shows the Dice values of the above models under different thresholds (used to binarize the mask). From these curves, we observe that BoxPolyp consistently outperforms other models, which proves its good capability for polyp segmentation.

\subsection{Visual Comparison}
\begin{figure}
  \centering
  \includegraphics[scale=0.48]{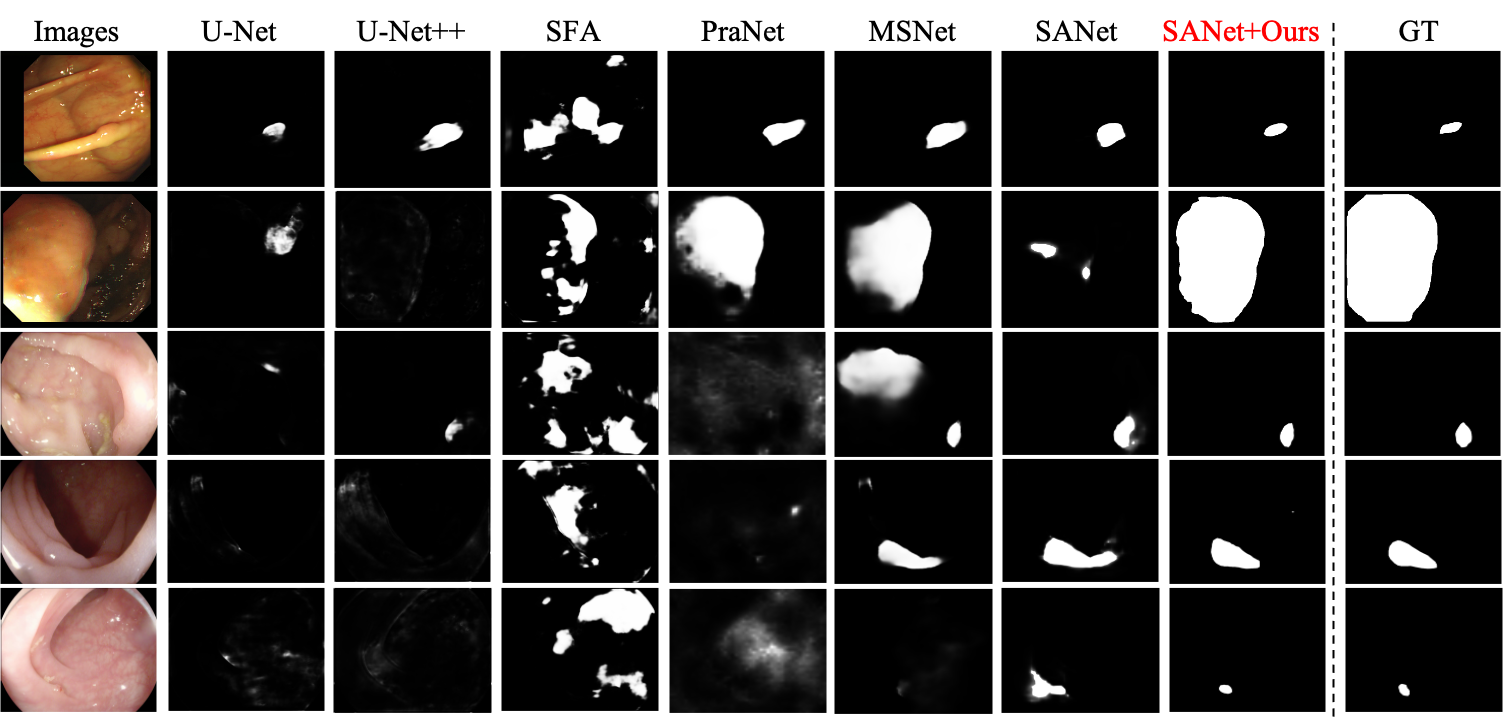}
  \caption{Visual comparison between the proposed method and six state-of-the-art ones.}
  \label{fig:visualization}
\end{figure}

Fig.~\ref{fig:visualization} visualizes some predictions of different models. Compared with other counterparts, our method not only clearly highlights the polyp regions but also suppresses the background noise. Even for challenging scenarios, our model still handles well and generates accurate segmentation mask.

\subsection{Ablation Study}
To investigate the importance of each component in BoxPolyp, the weighted average (wAVG) performace is adopted. We evaluate the model on both Res2Net~\cite{GaoCZZYT21} and PVT~\cite{wang2021pvtv2} for ablation studies. As shown in Table~\ref{tab:ablation}, all proposed modules are beneficial for the final predictions. Combining all these modules, our model achieves the new state-of-the-art performance.
\begin{table}[t]
  \caption{Ablation studies for BoxPolyp with different backbone networks.}
  \label{tab:ablation}
  \renewcommand\tabcolsep{10pt}
  \renewcommand\arraystretch{1.1}
  \centering
  \begin{tabular}{l|cc|cc}
    \hline
    \multicolumn{1}{c|}{\multirow{2}{*}{Settings}} & \multicolumn{2}{c|}{wAVG-Res2Net} & \multicolumn{2}{c}{wAVG-PVT}    \\
    \multicolumn{1}{c|}{}                       & mDice   & mIoU    & mDice   & mIoU    \\
    \hline
    SANet                                       & 0.794   & 0.714   & 0.833   & 0.760 \\
    SANet+FFS                                   & 0.839   & 0.757   & 0.848   & 0.772 \\
    SANet+FFS+IC                                & 0.846   & 0.771   & 0.851   & 0.776 \\
    \hline
  \end{tabular}
\end{table}

\section{Conclusion}
Limited by the size of the dataset, existing polyp segmentation models are vulnerable to noise and suffer from over-fitting.
For the first time, we leverage the cheap bounding box annotations to alleviate data shortage for a polyp segmentation task. Although coarse, these annotations can greatly improve the model generalization. It is achieved by the proposed FFS module and IC loss. In the future, we will explore the design of a weakly-supervised polyp segmentation model based on only bounding box annotations without masks. 

\section{Acknowledgement}
This work is supported by the Guangdong Provincial Key Laboratory of Big Data Computing, The Chinese University of Hong Kong, Shenzhen, by NSFC-Youth 61902335, by Key Area R\&D Program of Guangdong Province with
grant No.2018B030338001, by the National Key R\&D Program of China with grant No.2018YFB1800800, by Shenzhen Outstanding Talents Training Fund, by Guangdong Research Project No.2017ZT07X152, by Guangdong Regional Joint Fund-Key Projects 2019B1515120039, by the NSFC 61931024\&81922046, by helixon biotechnology company Fund and CCF-Tencent Open Fund.

\bibliographystyle{bibstyle}
\bibliography{bibliography}
\end{document}